\documentclass[final,3p,times,twocolumn]{elsarticle}

\usepackage{amssymb}
\usepackage{lineno}
\usepackage[numbers]{natbib}
\usepackage{epsfig}
\usepackage{graphicx}
\usepackage{amsmath}
\usepackage{amssymb}
\usepackage{pifont}% http://ctan.org/pkg/pifont
\usepackage{setspace}

\usepackage[utf8]{inputenc} % allow utf-8 input
\usepackage[T1]{fontenc}    % use 8-bit T1 fonts
\usepackage{booktabs}       % professional-quality tables
\usepackage{amsfonts}       % blackboard math symbols
\usepackage{nicefrac}       % compact symbols for 1/2, etc.
\usepackage{microtype}      % microtypography
\usepackage{color}
\usepackage{bigstrut}
\usepackage{hhline}% http://ctan.org/pkg/hhline
\usepackage{colortbl}
\usepackage{tabularx}
\usepackage{subcaption}
\usepackage{arydshln}
\usepackage{array,multirow}
\usepackage{xfrac}
\usepackage{bm}
\usepackage{soul}
\usepackage[export]{adjustbox}
\usepackage{graphicx}
\usepackage{verbatim}
\usepackage{mwe}
\usepackage[shortcuts]{extdash}
\usepackage{enumitem}
\usepackage{bbold}
\usepackage[normalem]{ulem}
\usepackage[dvipsnames]{xcolor}
\usepackage{bbm}
\usepackage{ulem}
\usepackage{float}
\usepackage{enumitem} % noindent for itemize
\usepackage{url}
\usepackage{subcaption}
\usepackage[justification=centering]{subcaption}
\usepackage{dsfont}
\usepackage{makecell} % new
\usepackage{svg}

\usepackage[ruled, lined, linesnumbered, commentsnumbered, longend]{algorithm2e}
\usepackage{algpseudocode}

\usepackage[pagebackref=false,breaklinks=true,colorlinks,bookmarks=false]{hyperref}

\newcommand{\ie}{\textit{i.e., }}
\newcommand{\eg}{\textit{e.g., }}
\newcommand{\cmark}{\ding{51}}%
\newcommand{\xmark}{\ding{55}} % new
\newcolumntype{L}[1]{>{\raggedright\let\newline\\\arraybackslash\hspace{0pt}}m{#1}}
\newcolumntype{R}[1]{>{\raggedleft\let\newline\\\arraybackslash\hspace{0pt}}m{#1}}
\newcolumntype{C}[1]{>{\centering\let\newline\\\arraybackslash\hspace{0pt}}m{#1}}
\definecolor{pastelblue}{rgb}{0.67, 1.0, 0.99}
\definecolor{pastelorange}{rgb}{1.0, 0.99, 0.78}
\definecolor{pastelyellow}{rgb}{0.60, 0.98, 0.60}

\newpageafter{title}
\newpageafter{author}
\newpageafter{abstract}

\journal{Pattern Recognition}

\begin{document}
\begin{frontmatter}



\title{Robust Pedestrian Detection via Constructing Versatile \\ Pedestrian Knowledge Bank}


\author[1]{Sungjune Park\fnref{cor1}}
\ead{sungjune-p@kaist.ac.kr}
\author[1]{Hyunjun Kim\fnref{cor1}}
\ead{kimhj709@kaist.ac.kr}
\author[1]{Yong Man Ro\corref{cor2}}
\ead{ymro@kaist.ac.kr}

\nonumnote{\textbf{This manuscript has been accepted for the publication in \textit{Pattern Recognition (2024)}.}}
\fntext[cor1]{Both authors contributed equally to this manuscript.}
\cortext[cor2]{Corresponding author.}

\affiliation[1]{
            organization={Image and Video Systems Lab., School of Electrical Engineering, Korea Advanced Institute of Science and Technology (KAIST)},
            city={Daejeon},
            postcode={34141},
            country={Republic of Korea}}

\begin{abstract}
Pedestrian detection is a crucial field of computer vision research which can be adopted in various real-world applications (\eg self-driving systems). However, despite noticeable evolution of pedestrian detection, pedestrian representations learned within a detection framework are usually limited to particular scene data in which they were trained. Therefore, in this paper, we propose a novel approach to construct versatile pedestrian knowledge bank containing representative pedestrian knowledge which can be applicable to various detection frameworks and adopted in diverse scenes. We extract generalized pedestrian knowledge from a large-scale pretrained model, and we curate them by quantizing most representative features and guiding them to be distinguishable from background scenes. Finally, we construct versatile pedestrian knowledge bank which is composed of such representations, and then we leverage it to complement and enhance pedestrian features within a pedestrian detection framework. Through comprehensive experiments, we validate the effectiveness of our method, demonstrating its versatility and outperforming state-of-the-art detection performances.
\end{abstract}



\end{frontmatter}

% \linenumbers

\section{Introduction} \label{sec:introduction}
Pedestrian detection has been studied actively as one of major applicable computer vision research \cite{pr_ped1, pr_ped2}. It can be widely applied to various real-world applications related to safety and security, such as self-driving and surveillance systems \cite{lowresolution, liu2023weakly, liu2024two}. Besides, as deep neural networks (DNNs) have emerged, pedestrian detection has also evolved rapidly showing noticeable performances \cite{pd11, wang2023prototype}.

%%%%%%%%%%%%%%%%%%%%%%%%%%%%%%%%% figure 1 %%%%%%%%%%%%%%%%%%%%%%%%%%%%%%%%%
\begin{figure*}[t]
\centering
	\includegraphics[width=0.95\textwidth]{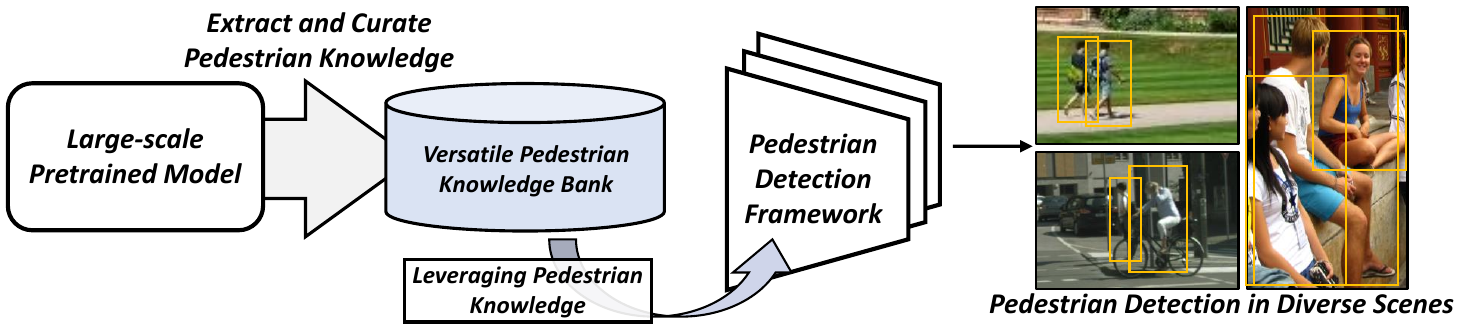}
    \caption{The overall concept of our approach. We extract generalized pedestrian knowledge from a large-scale pretrained model and curate them to be exemplary and task-compatible. The knowledge bank stores such knowledge, and it can be leveraged into various frameworks for robust pedestrian detection in diverse scene data.}
	\label{fig1}
\end{figure*}

%%%%%%%%%%%%%%%%%%%%%%%%%%%%%%%%% figure 1 %%%%%%%%%%%%%%%%%%%%%%%%%%%%%%%%%

Based on DNNs, pedestrian detection has tried to handle several real-world problems \cite{pr_ped2, pr_ped3, kim2021robust}. Liu \textit{et al.} \cite{ped4} proposed adaptive non-maximum suppression (adaptive-NMS) to apply suppression thresholds to each instance depending on their density scores. Rukhovich \textit{et al.} \cite{iterdet} designed an iterative pedestrian detection framework to refine object proposals by removing duplicated proposals in crowded scenes. Zheng \textit{et al.} \cite{progressive} presented a query-based pedestrian detection framework which accepts reliable queries and refines noisy queries progressively. Such methods mainly concentrated on designing delicate algorithms and frameworks to refine redundant object proposals. However, it has been discovered that pedestrian features learned within such frameworks are usually fitted to particular scenes used for training, thereby limiting their effectiveness in detecting pedestrians across diverse scenes \cite{elephant}.

Therefore, we are motivated by \textit{how can we acquire pedestrian representations that can be easily applicable to diverse scene data?}. In this paper, we propose a novel approach to construct versatile pedestrian knowledge bank consisting of pedestrian representations which can be utilized in various scenes. To this end, we benefit from a large-scale pretrained model which has generalized knowledge over numerous instances. We assemble numerous pedestrian samples, and then we feed them into a large-scale pretrained model (\textit{i.e.,} CLIP \cite{clip}). By doing so, we can extract the generalized knowledge of pedestrians. After that, we filter out representative pedestrian knowledge by utilizing vector quantization \cite{vector1}, and we guide them to be task-compatible with pedestrian detection. Since the representations from a large-scale pretrained model can be usually suboptimal to downstream tasks \cite{noisy1, vt-clip}, we make them distinguishable from various background scenes which are primary obstacles in pedestrian detection, that is, task-compatible. Then we store representative and task-compatible pedestrian features in versatile pedestrian knowledge bank, and it can be employed in various detection frameworks and diverse scene data. Finally, we leverage the knowledge bank to complement and enhance the pedestrian features within a pedestrian detection framework.

Figure \ref{fig1} illustrates the overall concept of our method. We construct versatile pedestrian knowledge bank with the help of a large-scale pretrained model. The knowledge bank can be exploited in various pedestrian detection frameworks enabling robust pedestrian detection on diverse scene data. Through extensive experiments with various detection frameworks and scene data, we corroborate the effectiveness and adaptability of our method showing state-of-the-art detection performances on four public pedestrian detection datasets, CrowdHuman, WiderPedestrian, CityPersons, and Caltech.

The main contributions can be summarized as follows:
\begin{itemize}
    \item We propose a novel method to acquire versatile pedestrian representations that are applicable to diverse frameworks and scene data. Based on generalized knowledge from a large-scale model, we obtain representative and task-compatible pedestrian knowledge features.
    \item We construct versatile pedestrian knowledge bank by storing the generalized and task-compatible knowledge, and then we leverage it to complement pedestrian features within various detection frameworks.
    \item We verify the effectiveness and versatility of our method with various detection frameworks and diverse datasets, achieving state-of-the-art performances on four pedestrian detection datasets.
\end{itemize}

\section{Related work}\label{sec:related}
\subsection{Object Detection}\label{sec:related-od}
As object detection has been considered as one of the fundamental computer vision tasks, it has been developed in various fronts. One of the main streams is region proposal based two stage object detection which is dominantly adopted in modern object detection frameworks (\textit{e.g.,} Cascade R-CNN \cite{cascade}). Recently, query based object detection has emerged \cite{sparse, deformable}. Sparse R-CNN \cite{sparse} adopts a query mechanism into Cascade R-CNN regarding each object query as an object candidate. D-DETR \cite{deformable} considers object detection as a set prediction problem and learns object queries with deformable convolutions in transformer architecture.

\subsection{Pedestrian Detection}
While pedestrian detection is regarded as critical real-world applications, it has also developed with sophisticated architectures and algorithms considering the characteristic of pedestrians \cite{bodla, progressive}. Several methods have been proposed with newly designed algorithms including non-maximum suppression (NMS) and training objectives. Bodla \textit{et al.} \cite{bodla} observed that, NMS is prone to remove neighboring objects. So, they modified the traditional NMS by adjusting detection scores depending on the overlap with the offset proposal box. Similarly, Liu \textit{et al.} \cite{ped4} proposed adaptive-NMS applying the threshold dynamically by predicting density scores for each instance. Huang \textit{et al.} \cite{ped3} designed representative region NMS (R$^2$NMS) to consider visible parts of pedestrians in NMS based on that visible parts usually suffer much less from overlap and occlusion. Tang \textit{et al.} \cite{pd11} presented OTP-NMS to predict optimal threshold of NMS by considering visibility and classification score for each proposal. It tried to mitigate high overlaps in the crowded scenes by regulating NMS threshold. Zhang \textit{et al.} \cite{ped2} introduced a variational pedestrian detection framework by formulating object proposals as latent variables and considering detection as a variational inference problem. Zhang \textit{et al.} \cite{attribute} considered pedestrians' attribute features and introduced to a pedestrian oriented attribute features by encoding an attribute map with center, offset, and scale maps. Then they designed attribute-NMS to filter out false positives in crowded scenes. Zheng \textit{et al.} \cite{progressive} proposed an iterative query based pedestrian detector which repeats accept and reject steps iteratively. It accepts queries with high scores, and a relation information extractor updates the residual queries to be refined.

However, it is discovered that the learned pedestrian features from such frameworks are usually fitted to the particular scenes in the training data, limiting their ability to detect pedestrians in diverse scenes \cite{elephant}. Therefore, we are motivated that \textit{how to obtain versatile pedestrian features which can be applicable on diverse scenes}.

%%%%%%%%%%%%%%%%%%%%%%%%%%%%%%%%% figure 2 %%%%%%%%%%%%%%%%%%%%%%%%%%%%%%%%%
\begin{figure*}[t]
\centering
	\includegraphics[width=0.999\textwidth]{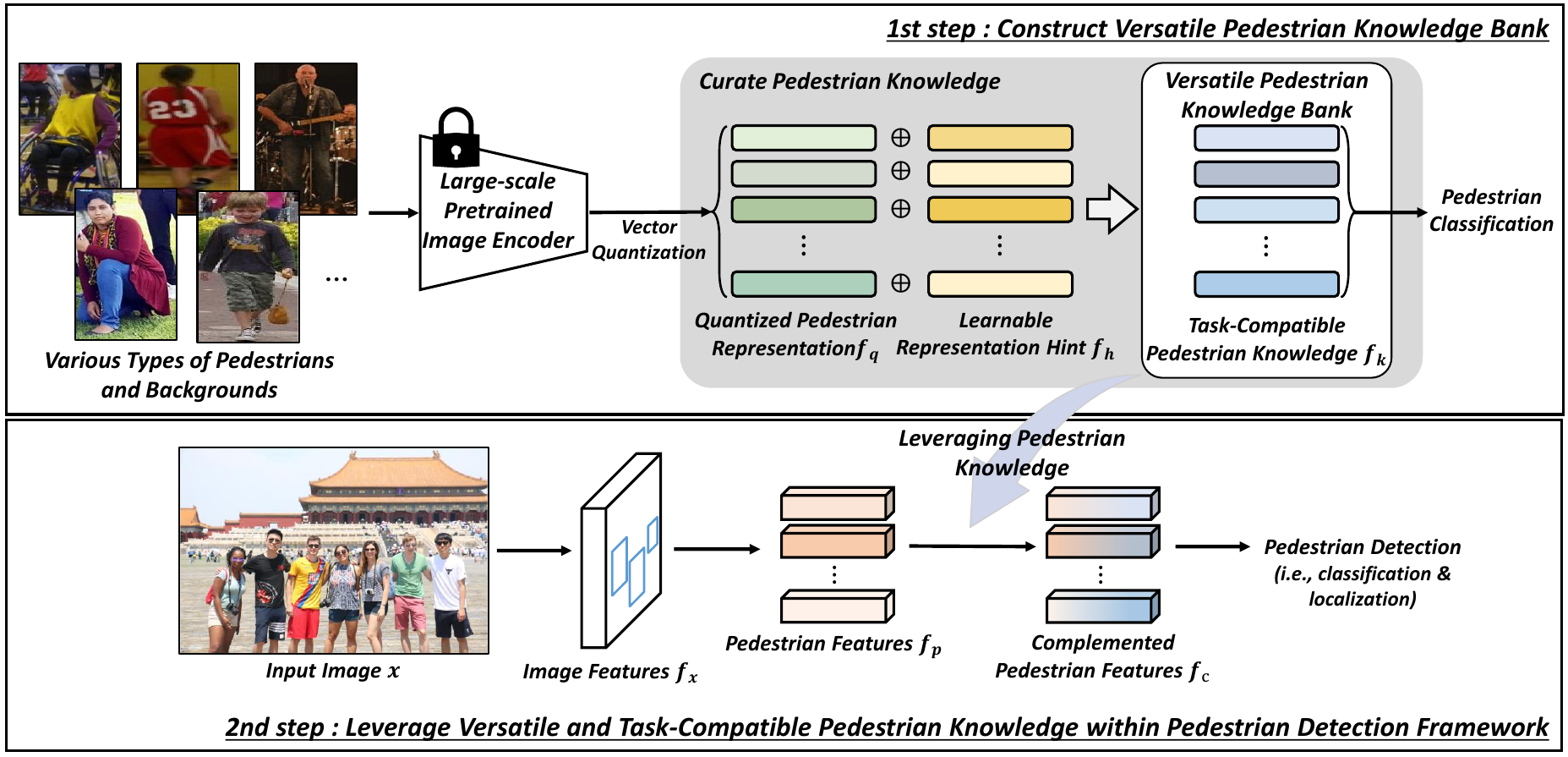}
    \caption{The overall steps designed in the proposed approach. At the first step, we extract the knowledge embeddings of various instances from a large-scale pretrained image encoder. We quantize the most representative $\boldsymbol{f_q}$ and make them task-relevant by placing $\boldsymbol{f_h}$. Then we obtain task-compatible knowledge features $\boldsymbol{f_k}$. At the second step, we leverage $\boldsymbol{f_k}$ within a pedestrian detection framework.}
	\label{fig2}
\end{figure*}
%%%%%%%%%%%%%%%%%%%%%%%%%%%%%%%%% figure 2 %%%%%%%%%%%%%%%%%%%%%%%%%%%%%%%%%

\subsection{Generalized Knowledge in Large-scale Pretrained Models}
In recent years, several large-scale pretrained models have been introduced, and they are usually trained with massive image and text paired data via contrastive learning. CLIP \cite{clip} is trained on 400 million web-crawled image-text pairs, and it demonstrates a noticeable transfer ability on zero-shot inference tasks \cite{zero6, zero7}. Kobs \textit{et al.} \cite{zero6} proposed a language-guided zero-shot deep metric learning method by utilizing CLIP as general purpose image and language feature extractors. Sanghi \textit{et al.} \cite{zero7} adopted CLIP text encoder to acquire query text features, and then conditioned the flow network with the text features to obtain shape embedding for zero-shot text-to-shape generation. Gu \textit{et al.} \cite{zero9} used CLIP image encoder to guide object region features to be similar with CLIP image features, and then CLIP text encoder is utilized as an object classifier. Also, Ci \textit{et al.} \cite{unihcp} proposed a unified model for human-centric perceptions, named UniHCP, which is jointly trained with large-scale database (\textit{i.e.,} 33 datasets for 5 different tasks including pedestrian detection). By doing so, it could learn general human-centric knowledge which can help to boost the performance of downstream pedestrian detection.

Although large-scale pretrained models have shown a notable capability in encoding generalized representations across numerous instances, several works have observed that: they are usually insufficient to distinguish fine-grained characteristics of instances, and their representations could be suboptimal on downstream tasks because of semantic gap between tasks \cite{noisy1, vt-clip}. Therefore, in this paper, we try to mitigate such a gap while obtaining versatile pedestrian knowledge bank by placing a learnable representation hint, so that it contains task-compatible knowledge features.

\section{Proposed Method}
Figure \ref{fig2} shows the overall process of the proposed method. The first step (\textit{upper figure}) is to construct versatile pedestrian knowledge bank. At this step, we extract the generalized pedestrian embeddings through a large-scale model, and then we curate them by quantizing the most representatives. They are also guided to be well-separated from background scenes to be task-compatible with pedestrian detection. By storing them, we construct versatile pedestrian knowledge bank. The second step (\textit{lower figure}) is to leverage task-compatible knowledge features within a pedestrian detection framework. The pedestrian features within a detection framework can be complemented by distinctive pedestrian knowledge, and then the complemented pedestrian features are used to perform final pedestrian detection (\ie classification and localization). Note that the first step is included in the training phase only, not the testing phase. More details for each step are elaborated in the following subsections.

\subsection{How to Construct Versatile Pedestrian Knowledge Bank}
\label{sect3a}
The purpose of the first step is to construct versatile pedestrian knowledge bank consisting of $N$ knowledge features. As shown in Figure \ref{fig2}, we prepare an image set which contains various types of pedestrian and background images. Using CrowdHuman \cite{crowdhuman}, we randomly select and crop pedestrian instances to include various representations of pedestrians, and we also collect background regions that are not overlapped with pedestrian instances. When we feed pedestrian cropped images into CLIP image encoder, we can extract the generalized pedestrian embeddings. Then we curate the extracted features to construct versatile pedestrian knowledge bank. We quantize them by performing k-means clustering to obtain the most representative pedestrian features $\boldsymbol{f_q}=\{\boldsymbol{f^i_q}\}^N_{i=1} \in \mathbb{R}^{N \times d}$ where $N$ and $d$ are the number of quantized features and the channel dimension, respectively.

Not using $\boldsymbol{f_q}$ directly, we further guide them to be distinct from various non-object background scenes, and then, we obtain knowledge features $\boldsymbol{f_k}=\{\boldsymbol{f^i_k}\}^N_{i=1} \in \mathbb{R}^{N \times d}$ which is compatible with pedestrian detection task. To this end, we place learnable representation hint $\boldsymbol{f_h}=\{\boldsymbol{f^i_h}\}^N_{i=1} \in \mathbb{R}^{N \times d}$ to be added with each of $\boldsymbol{f_q}$ (described as $\oplus$ in Figure \ref{fig2}). Due to $\boldsymbol{f_h}$, $\boldsymbol{f_q}$ can be transformed to be more related to our purpose separating pedestrians and backgrounds well. When an input pedestrian instance comes in, the image encoder extracts pedestrian feature embedding $\boldsymbol{p} \in \mathbb{R}^{1 \times d}$. Then we perform vector quantization \cite{vector1} to find the closest representation among $N$ number of quantized $\boldsymbol{f_q}$ as follows,
\begin{equation}
    n = \underset{i \in N}{argmax} \ \boldsymbol{f^i_q} \boldsymbol{p}^T.
    \label{eq1}
\end{equation}
Here, $n$ is the $n$-th index among $N$ quantized $\boldsymbol{f_q}$, so that $\boldsymbol{f^n_q}$ is the most similar pedestrian representation with the input embedding $\boldsymbol{p}$. After that, we obtain task-compatible knowledge features of $\boldsymbol{f_k}$ as follows,
\begin{equation}
    \boldsymbol{f^n_k} = \boldsymbol{f^n_q} \oplus \boldsymbol{f^n_h},
    \label{eq2}
\end{equation}
where $\oplus$ is element-wise summation. The obtained $\boldsymbol{f_k}$ is called task-compatible knowledge features, because we guide them to solve pedestrian classification problem being more relevant with pedestrian detection. For pedestrian classification, we also conduct the above-mentioned process with background input images at the same time. Then we conduct pedestrian classification with the obtained pedestrian and background features by utilizing two fully connected (FC) layers. Consequently, learnable representation hint $\boldsymbol{f_h}$ is updated to adjust $\boldsymbol{f_q}$ and to acquire task-compatible $\boldsymbol{f_k}$.

Based on generalized knowledge from a large-scale pretrained model, we curate pedestrian representations to be distinctive from various non-object background scenes (\textit{task-compatible}). After storing them in versatile pedestrian knowledge bank, and we can leverage them into various pedestrian detectors. More details are described in the following section.

\subsection{How to Leverage Versatile Pedestrian Knowledge}
In this subsection, we describe how to leverage versatile pedestrian knowledge from the bank into a pedestrian detection framework. We explain it separately depending on two types of detection framework, region proposal based two stage detection (\textit{e.g.,} Cascade R-CNN) and query based detection (\textit{e.g.,} D-DETR).

\subsubsection{Region Proposal Based Two Stage Detection}
\label{sect321}
We firstly explain the overall flow of exploiting the knowledge in a detection framework (the second step in Figure \ref{fig2}). The image features $\boldsymbol{f_x} \in \mathbb{R}^{H \times W \times C}$ are obtained from an input image $\boldsymbol{x}$, where $H$, $W$, and $C$ are height, width, and channel dimension of the feature map. In region proposal based two stage detection framework, region proposal network (RPN) predicts region candidates which seem to be pedestrians by taking $\boldsymbol{f_x}$. Then region pooling is conducted to obtain proposal features. Such proposal features are considered to represent pedestrian candidates in $\boldsymbol{f_x}$, called pedestrian features $\boldsymbol{f_p}=\{\boldsymbol{f^i_p}\}^M_{i=1}$, where $\boldsymbol{f^i_p} \in \mathbb{R}^{h \times w \times c}$ and $M$ is the number of pedestrian features. While we adopt cross-attention architecture to leverage the knowledge features, $\boldsymbol{f_p}$ is taken as input query features, and task-compatible pedestrian knowledge features $\boldsymbol{f_k}$ from the versatile bank are taken as key and value features. So that, the pedestrian features are complemented, and we can acquire $\boldsymbol{f_c}=\{\boldsymbol{f^i_c}\}^M_{i=1}$. The dimension of each $\boldsymbol{f^i_c}$ is $\mathbb{R}^{h \times w \times c}$. Finally, pedestrian detection (\ie classification and localization) is conducted with $\boldsymbol{f_c}$ to find where pedestrians are located in the input image $\boldsymbol{x}$.

%%%%%%%%%%%%%%%%%%%%%%%%%%%%%%%%% figure 3 %%%%%%%%%%%%%%%%%%%%%%%%%%%%%%%%%
\begin{figure}[t]
\centering
	\includegraphics[width=0.999\columnwidth]{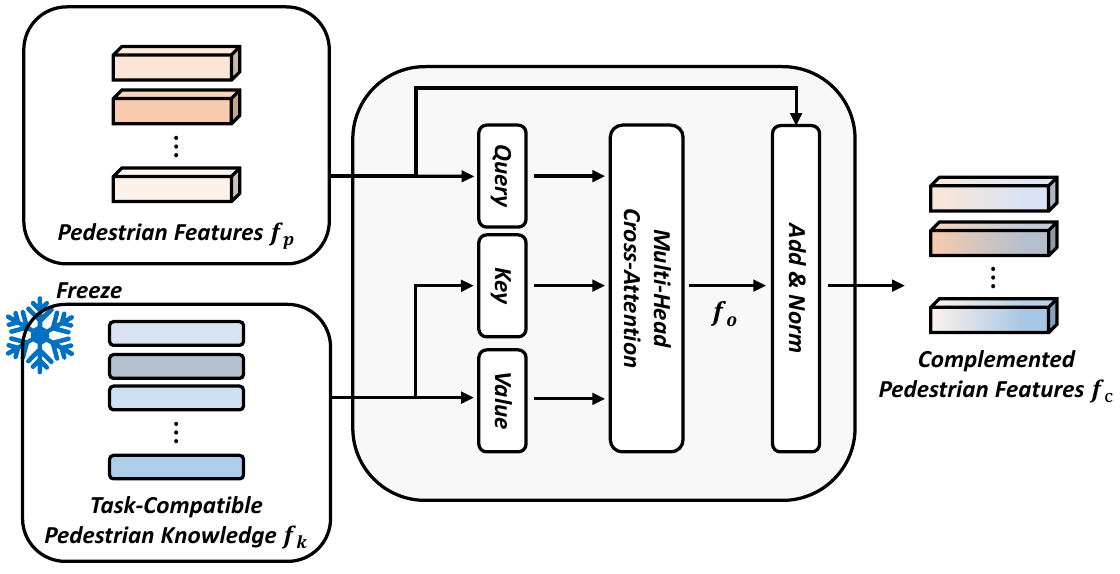}
    \caption{The overview of leveraging task-compatible pedestrian knowledge $\boldsymbol{f_k}$. When pedestrian features $\boldsymbol{f_p}$ come in as query features, $\boldsymbol{f_k}$ functions as key and value features. So, $\boldsymbol{f_p}$ can refer to $\boldsymbol{f_k}$, distinguishable features from the bank, then the complemented pedestrian features $\boldsymbol{f_c}$ can be obtained.}
	\label{fig3}
\end{figure}

%%%%%%%%%%%%%%%%%%%%%%%%%%%%%%%%% figure 3 %%%%%%%%%%%%%%%%%%%%%%%%%%%%%%%%%

The details are described in Figure \ref{fig3}. We give an explanation with the $i\text{-}th$ pedestrian feature $\boldsymbol{f^i_p} \in \mathbb{R}^{h \times w \times c}$ for simplicity. When $\boldsymbol{f^i_p}$ comes in, it is flattened to $\boldsymbol{\bar{f^i_p}} \in \mathbb{R}^{hw \times c}$. $\boldsymbol{\bar{f^i_p}}$ is projected to query features $\boldsymbol{Q} \in \mathbb{R}^{hw \times d_m}$ as follows,
\begin{equation}
    \boldsymbol{Q} = \boldsymbol{\bar{f^i_p}} \boldsymbol{W_Q},
    \label{eq3}
\end{equation}
where $\boldsymbol{W_Q}$ denotes a query projection matrix of $\mathbb{R}^{c \times d_m}$. Next, we associate pedestrian query features (encoded within a pedestrian detection framework) with the task-compatible knowledge features $\boldsymbol{f_k}$ from the bank. So, key and value features $\boldsymbol{K}, \boldsymbol{V} \in \mathbb{R}^{N \times d_m}$ are obtained by
\begin{equation}
    \begin{aligned}
        \boldsymbol{K} = \boldsymbol{f_k} \boldsymbol{W_K}, \qquad\qquad \boldsymbol{V} = \boldsymbol{f_k} \boldsymbol{W_V},
    \end{aligned}
    \label{eq4}
\end{equation}
where $\boldsymbol{W_K}$ and $\boldsymbol{W_V}$ are key and value projection matrices of $\mathbb{R}^{d \times d_m}$, respectively. We calculate feature similarity between $\boldsymbol{Q}$ and $\boldsymbol{K}$ and obtain feature association vector $\boldsymbol{A} \in \mathbb{R}^{hw \times N}$ as follows,
\begin{equation}
    \boldsymbol{A} = softmax(\frac{\boldsymbol{Q} \boldsymbol{K}^T}{\sqrt{d_m}}),
    \label{eq5}
\end{equation}
where $d_m$ is the dimension of $\boldsymbol{Q}$ and $\boldsymbol{K}$. Based on the obtained $\boldsymbol{A}$, we aggregate value features $\boldsymbol{V}$ and obtain output features $\boldsymbol{f_o} \in \mathbb{R}^{hw \times c}$ by
\begin{equation}
    \boldsymbol{f_o} = \boldsymbol{A} \boldsymbol{V} \boldsymbol{W_O},
    \label{eq6}
\end{equation}
where $\boldsymbol{W_O} \in \mathbb{R}^{d_m \times c}$ denotes output projection matrix. The obtained $\boldsymbol{f_o}$ is then added with $\boldsymbol{f_p}$ by residual connection, and layer normalization is performed to generate $\boldsymbol{\bar{f^i_c}} \in \mathbb{R}^{hw \times c}$ which can be reshaped into the complemented pedestrian features $\boldsymbol{f^i_c} \in \mathbb{R}^{h \times w \times c}$.

Furthermore, we perform such a cross-attention in parallel to leverage the knowledge features stored in the bank jointly (\ie multi-head cross-attention). To this end, we perform $l$ number of different projection matrices as follows,
\begin{equation}
    \begin{aligned}
        \boldsymbol{f_o} = Concat(\boldsymbol{f_{o,1}}, \boldsymbol{f_{o,2}}, \cdots, \boldsymbol{f_{o,l}}) \boldsymbol{W_O'}, \qquad \\ where \boldsymbol{f_{o,j}} = \boldsymbol{A_j} \boldsymbol{V_j}.
    \end{aligned}
    \label{eq7}
\end{equation}

\noindent
By doing so, we can leverage versatile and task-compatible pedestrian knowledge features in order to complement pedestrian features within a pedestrian detection framework, helping them more distinctive from various backgrounds. To validate it, we adopt the proposed pedestrian knowledge bank in various pedestrian detection frameworks, and we validate it on diverse scene data using four pedestrian detection benchmarks.

For the training of detection framework, we employ traditional pedestrian detection training objectives. Usually, in region-proposal based two stage detection, cross-entropy and smooth-l1 losses are used for RPN and last detection classification and localization, respectively.

\subsubsection{Query Based Detection}
Different from region proposal based two stage detection, query based detection utilizes a set of object queries. We consider $\boldsymbol{f_p}=\{\boldsymbol{f^i_p}\}^M_{i=1}$ as object query features (not object proposal features), where $M$ is the predefined number of object queries. And each of them is $\boldsymbol{f^i_p} \in \mathbb{R}^{1 \times c}$, so that it is not required to be flattened when being projected to query features. Except that, the way to leverage the knowledge bank is totally same with that of region proposal based two stage detection as described in Section \ref{sect321}. For the training of detection framework, we use training objectives commonly used in query based detectors: focal loss for classification, and L1 and GIoU losses for localization.

For the better understanding, Algorithm \ref{alg1} describes the overall procedure of the proposed method including first and second steps regardless of the type of detection frameworks.

%%%%%%%%%%%%%%%%%%%%%%%%%%%%%%% Alg 1 %%%%%%%%%%%%%%%%%%%%%%%%%%%%%%%
\newcommand\mycommfont[1]{\footnotesize\ttfamily{#1}}
\SetCommentSty{mycommfont}
\begin{algorithm}[hbt!]
    \SetKwInOut{KwIn}{Input}
    \SetKwInOut{KwOut}{Output}
    \KwIn{\small{Image set of various instances $\boldsymbol{S}$; Input image $\boldsymbol{x}$.}}
    \KwOut{\small{Versatile pedestrian knowledge bank $\boldsymbol{f_k}=\{\boldsymbol{f^i_k}\}^N_{i=1}$ (1st step); Complemented pedestrian features $\boldsymbol{f_c}=\{\boldsymbol{f^i_c}\}^M_{i=1}$ (2nd step).}}
    \BlankLine
    \small\tcc{1st step: Construct $\boldsymbol{f_k}$.}
    \small\tcp{Extract quantized representations.}
    \small{$\boldsymbol{P} \leftarrow Enc(\boldsymbol{S})$}    \Comment{\footnotesize{\textcolor{gray}{$Enc(\cdot)$: CLIP image encoder.}}} \\
    \small{$\boldsymbol{f_q} \leftarrow VQ(\boldsymbol{P})$}    \Comment{\footnotesize{\textcolor{gray}{$VQ(\cdot)$: Vector quantization.}}} \\
    \small\tcp{Learn task-compatible $\boldsymbol{f_h}=\{\boldsymbol{f^i_h}\}^N_{i=1}$.}
    \small\While{not converging}{
        \small{$\boldsymbol{p} \leftarrow$ random sampled from $\boldsymbol{P}$} \\
        \small{$n \leftarrow \underset{i \in N}{argmax} \ \boldsymbol{f^i_q} \boldsymbol{p}^{\top}$} \\
        \small{$\boldsymbol{f^n_k} \leftarrow \boldsymbol{f^n_q} \oplus \boldsymbol{f^n_h}$} \\
        \small{Update $\boldsymbol{f^n_h}$ by solving binary pedestrian classification with $FC(\boldsymbol{f^n_k})$} \\
    }
    \small{$\boldsymbol{f_k} \leftarrow \boldsymbol{f_q} \oplus \boldsymbol{f_h}$} \\
    \small\KwRet{$\boldsymbol{f_k}$}
    \BlankLine
    
    \small\tcc{2nd step: Leverage $\boldsymbol{f_k}$ within detection framework.}
    \small\tcp{Obtain object proposal/query features $\boldsymbol{f_p}=\{\boldsymbol{f^i_p}\}^M_{i=1}$.}
    \small{$\boldsymbol{f_x} \leftarrow FE(\boldsymbol{x})$}   \Comment{\footnotesize{\textcolor{gray}{$FE(\cdot)$: Feature extractor.}}} \\
    \small{$\boldsymbol{f_p} \leftarrow$ Object proposal/query features for given $\boldsymbol{f_x}$} \\
    \small\tcp{Leverage $\boldsymbol{f_k}$ and obtain $\boldsymbol{f_c}$.}
    % f_p = M X c / f_k = N X d / f_c = M X c
    \small{$\boldsymbol{f_c} \leftarrow \boldsymbol{f_p} + MHCA(\boldsymbol{f_p}, \boldsymbol{f_k}, \boldsymbol{f_k})$} \Comment{\footnotesize{\textcolor{gray}{$MHCA(\cdot)$: Multi-head cross-attention.}}} \\
    \small\KwRet{$\boldsymbol{f_c}$} \\
    \BlankLine
    \small\tcc{Obtain final pedestrian detection results with $\boldsymbol{f_c}$.}
    \caption{Procedure of the proposed method.}
    \label{alg1}
\end{algorithm}
%%%%%%%%%%%%%%%%%%%%%%%%%%%%%%% Alg 1 %%%%%%%%%%%%%%%%%%%%%%%%%%%%%%%

\section{Experiments}
In this section, we describe the experimental settings including pedestrian detection datasets and implementation details. Then we show extensive experimental results to validate the effectiveness of our method.

\subsection{Experimental Settings}
\subsubsection{Pedestrian Detection Datasets}
\noindent \textbf{CrowdHuman} \cite{crowdhuman} is one of large pedestrian detection datasets composed of web-crawled images. It consists of 15,000, 4,370, and 5,000 images for each training, validation, and test sets, respectively. There are about 340,000 pedestrian instances in the training set, and there are three kinds of pedestrian annotations: head, visible-body, and full-body bounding boxes. For fair comparison with other methods, we use training and validation subsets with full-body annotations. We randomly select pedestrian instances from the training set of CrowdHuman and crop them to construct versatile pedestrian knowledge bank. Please note that we do not use any other benchmarks to build the proposed knowledge bank.

\noindent \textbf{WiderPedestrian} \cite{wider} is another large pedestrian detection dataset, composed of the images captured at surveillance and driving environments in urban city. There are 11,500 training images, 5,000 validation images, and 3,500 testing images. Since ground-truth annotations for testing images is inaccessible, we use the validation set for evaluation.

\noindent \textbf{CityPersons} \cite{city} is a widely known pedestrian detection dataset. It includes image data of driving scenes collected in 18 cities under various weather conditions, and it is composed of training, validation, and test subsets, respectively. We use the validation subset for fair evaluation. The number of pedestrians in the training set is about 20,000, and the resolution of images is 1024 $\times$ 2048. We conduct training and evaluation using 2,975 training and 500 validation images. We utilize `Reasonable' subset which encompasses a height range of [$0.50$, $inf$] and a visibility ratio of [$0.65$, $inf$] for both training and evaluation. Following the existing methods \cite{progressive, ped2}, we also use full-body annotations for training and evaluation.

\noindent \textbf{Caltech} \cite{caltech} is another traditional pedestrian detection dataset containing images from driving scenes. It is collected from the video recorded in Los Angeles, and it includes about 14,000 pedestrian instances in the training data. We employ 42,782 images augmented by 10 folds for training along with 4,024 testing images for evaluation. We also utilize `Reasonable' subset with the same height and visibility range constraints as in CityPersons dataset.

%%%%%%%%%%%%%%%%%%%%%%%%%%%%%%%%%%% Table 1 %%%%%%%%%%%%%%%%%%%%%%%%%%%%%%%%%%%
% \rule{0pt}{9pt}   
\begin{table*}[t]
    \caption{The comparison with existing pedestrian detection methods on CrowdHuman. `\# Queries' denotes the number of queries. For evaluation metrics, we adopt AP and MR$^{-2}$. The higher AP means the better performance, and vice versa for MR$^{-2}$.}
	\centering
	\begin{center}
		\renewcommand{\tabcolsep}{12.0mm}
		\resizebox{0.99\linewidth}{!}
		{
			\begin{tabular}{cccc}
				\Xhline{3\arrayrulewidth}
                    \rule{0pt}{11pt}
				\bf Method   & \bf \# Queries & \bf AP ($\uparrow$)& \bf MR$^{-2}$ ($\downarrow$) \\ \hline
                    \rule{0pt}{11pt}
                    Cascade R-CNN (CVPR'18) \cite{cascade} & - & 85.6 & 43.0 \\
                    FPN \cite{fpn}+SoftNMS (ICCV'17) \cite{bodla} & - & 88.2 & 42.9 \\
                    UniHCP (CVPR'23) \cite{unihcp} & 900 & 90.0 & 46.6 \\
                    Sparse R-CNN (CVPR'21) \cite{sparse} & 500 & 90.7 & 44.7 \\
                    OTP-NMS (TIP'23) \cite{pd11} & - & 90.9 & 41.2 \\
                    OPL (CVPR'23) \cite{cvpr23_1} & - & 91.0 & 44.9 \\
                    Sparse R-CNN (CVPR'21) \cite{sparse} & 750 & 91.3 & 44.8 \\
                    D-DETR (arXiv'20) \cite{deformable} & 1000 & 91.5 & 43.7 \\
                    E2EDET (Sparse R-CNN) (CVPR'22) \cite{progressive} & 500 & 92.0 & 41.4 \\
                    E2EDET (D-DETR) (CVPR'22) \cite{progressive} & 1000 & 92.1 & 41.5 \\
                    DRFG+PMIP (PR'22) \cite{ped5} & - & 92.2 & 41.8 \\
                    UniHCP-FT (CVPR'23) \cite{unihcp} & 900 & 92.5 & 41.6 \\ \hline 
                    \rule{0pt}{11pt}
                    \bf Ours (Cascade R-CNN) & \bf - & \bf 89.0 & \bf 41.7 \\
                    \bf Ours (Sparse R-CNN) & \bf 500 & \bf 92.4 & \bf 42.1 \\
                    \bf Ours (Sparse R-CNN) & \bf 750 & \bf 92.7 & \bf 42.1 \\
                    \bf Ours (D-DETR) & \bf 1000 & \bf 94.2 & \bf 38.2 \\ \Xhline{3\arrayrulewidth}
			\end{tabular}
		}
	\end{center}
	\label{tab1} 
\end{table*}
%%%%%%%%%%%%%%%%%%%%%%%%%%%%%%%%%%% Table 1 %%%%%%%%%%%%%%%%%%%%%%%%%%%%%%%%%%%

\subsubsection{Implementation Details}
For the first step, we adopted CLIP ViT-B/32 image encoder which is widely used in previous works \cite{zero7, zero9}. The dimension of the extracted pedestrian embedding $\boldsymbol{p}$ is 512 ($d=512$). When training $\boldsymbol{f_h}$ with pedestrian classification, we employed binary cross entropy (BCE) loss using stochastic gradient descent (SGD) optimizer with learning rate $0.1$. In default, we store 50 versatile and task-compatible pedestrian knowledge features in the bank ($N=50$). For the second step, the dimension $d_m$ is set as $64$, and the number of parallel leveraging module is 8 ($l=8$). We adopted the proposed method with three pedestrian detection frameworks, Cascade R-CNN \cite{caltech}, Sparse R-CNN \cite{sparse}, and D-DETR \cite{deformable}. We adhered to the training details for each framework following \cite{elephant, progressive, deformable}. While they have a cascade architecture consisting of 6 decoding stages, we leveraged the knowledge bank at the last 6th stage.

\subsection{Comparison with Existing Pedestrian Detection Methods}
In this subsection, we compare the detection performance of our method with the existing methods. We use CrowdHuman \cite{crowdhuman} for the evaluation which is widely known as pedestrian detection dataset in crowded scenes. Average precision (AP) is used for the evaluation metric, mostly adopted in CrowdHuman \cite{progressive}. We also adopt log-averaged miss rate (MR$^{-2}$) with false positive per image (FPPI) in the range of [$10^{-2}$, $10^0$] following \cite{pd11}. `\# Queries' refers to the number of queries in query-based detection frameworks. Table \ref{tab1} describes the comparison results. For the experiments, we employ three detection frameworks to assess the versatility of our method, and it shows that the proposed method can obtain remarkable performance improvements with the various detection frameworks. For example, for Cascade R-CNN \cite{cascade}, one of the traditional two-stage detectors, the proposed method obtains 3.4 AP performance gain, from 85.6 AP to 89.0 AP. When our method is applied into query-based Sparse R-CNN \cite{sparse}, it also obtains large performance improvements. Furthermore, when adopting D-DETR \cite{deformable} with the proposed knowledge bank, it shows state-of-the-art pedestrian detection performance outperforming the other existing methods. From the experiments, we corroborate that the proposed method is so adaptable to various frameworks that it can boost the detection performance with large margins.

\subsection{Effectiveness on Diverse Scenes}
In this subsection, we validate the effectiveness regarding the applicability of the versatile pedestrian knowledge bank to diverse scene data. Please note that, the knowledge bank is constructed with pedestrian instances from the training data of CrowdHuman \cite{crowdhuman} only, without using other pedestrian detection benchmarks. To validate it on driving scene data, we conduct the experiments by using CityPersons \cite{city} and Caltech \cite{caltech}. For the experiments, we adopt MR$^{-2}$ as the evaluation metric which is mainly adopted evaluation metrics for two benchmarks \cite{progressive, elephant}, and we use Sparse R-CNN \cite{sparse} as the baseline detection framework. Table \ref{tab2} shows the experimental results on both benchmarks. For CityPersons \cite{city} (top table), the proposed method achieves 7.6 MR$^{-2}$ outperforming other existing methods. Also, for Caltech \cite{caltech}, another benchmark mainly collected in driving scene, our method also obtains 2.7 MR$^{-2}$. The experiments demonstrate the effectiveness of the proposed method on driving scene data.

%%%%%%%%%%%%%%%%%%%%%%%%%%%%%%%%%%% Table 2 %%%%%%%%%%%%%%%%%%%%%%%%%%%%%%%%%%%
% \rule{0pt}{9pt}   
\begin{table*}[t]
	\caption{The effectiveness of our method on driving scenes (CityPersons \cite{city} and Caltech \cite{caltech}). `\#Q' means the number of queries, and we adopt miss rate (MR$^{-2}$) as evaluation metric. The lower MR$^{-2}$ means the better performance.}
	\centering
	\begin{center}
		\renewcommand{\tabcolsep}{6.0mm}
		\resizebox{0.999\linewidth}{!}
		{
			\begin{tabular}{ccc||ccc}
                \Xhline{3\arrayrulewidth}
                \multicolumn{3}{c||}{\textit{CityPersons} \cite{city}} & \multicolumn{3}{c}{\textit{Caltech} \cite{caltech}} \\
                \bf Method & \bf \#Q & \bf MR$^{-2}$($\downarrow$) & \bf Method & \bf \#Q & \bf MR$^{-2}$($\downarrow$) \\ \hline
                \rule{0pt}{11pt}
                Cascade R-CNN (CVPR'18) \cite{cascade} & - & 11.2 & ALFNET (ECCV'18) \cite{alfnet} & - & 6.1 \\
                Sparse R-CNN (CVPR'21) \cite{sparse} & 500 & 10.0 & OPL (CVPR'23) \cite{cvpr23_1} & - & 5.2 \\
                FPN \cite{fpn}+SoftNMS (ICCV'17) \cite{bodla} & -  & 9.9 & CSP (PR'22) \cite{csp} & - & 4.5 \\
                OTP-NMS (TIP'23) \cite{pd11} & - & 9.5 & BGCNet (ACM MM'20) \cite{bgc} & - & 4.1 \\
                D-DETR (arXiv'20) \cite{deformable} & 500 & 9.4 & Sparse R-CNN (CVPR'21) \cite{sparse} & 500 & 3.1 \\
                E2EDET (CVPR'22) \cite{progressive} & 500 & 7.8 & E2EDET (CVPR'22) \cite{progressive} & 500 & 2.9 \\ \hline
                \rule{0pt}{11pt}
                \bf Ours & \bf 500 & \bf 7.6 & \bf Ours & \bf 500 & \bf 2.7 \\ \Xhline{3\arrayrulewidth}
\end{tabular}
		}
	\end{center}
	\label{tab2} 
\end{table*}
%%%%%%%%%%%%%%%%%%%%%%%%%%%%%%%%%%% Table 2 %%%%%%%%%%%%%%%%%%%%%%%%%%%%%%%%%%%

Furthermore, we also perform the experiment on the surveillance scene data by using WiderPedestrian benchmark \cite{wider} which is composed of driving and surveillance scene data together. Since the testing annotations are not available for now, we adopt the validation data set to evaluate and compare the existing methods following WIDER Face and Pedestrian Challenge 2018 \cite{wider}. The experimental results are shown in Table \ref{tab3}, and it corroborates that our method is also applicable on the surveillance scene data.

\subsection{Semantic Information in Pedestrian Knowledge Bank}
\begin{figure*}[t]
\centering
    \includegraphics[width=0.95\textwidth]{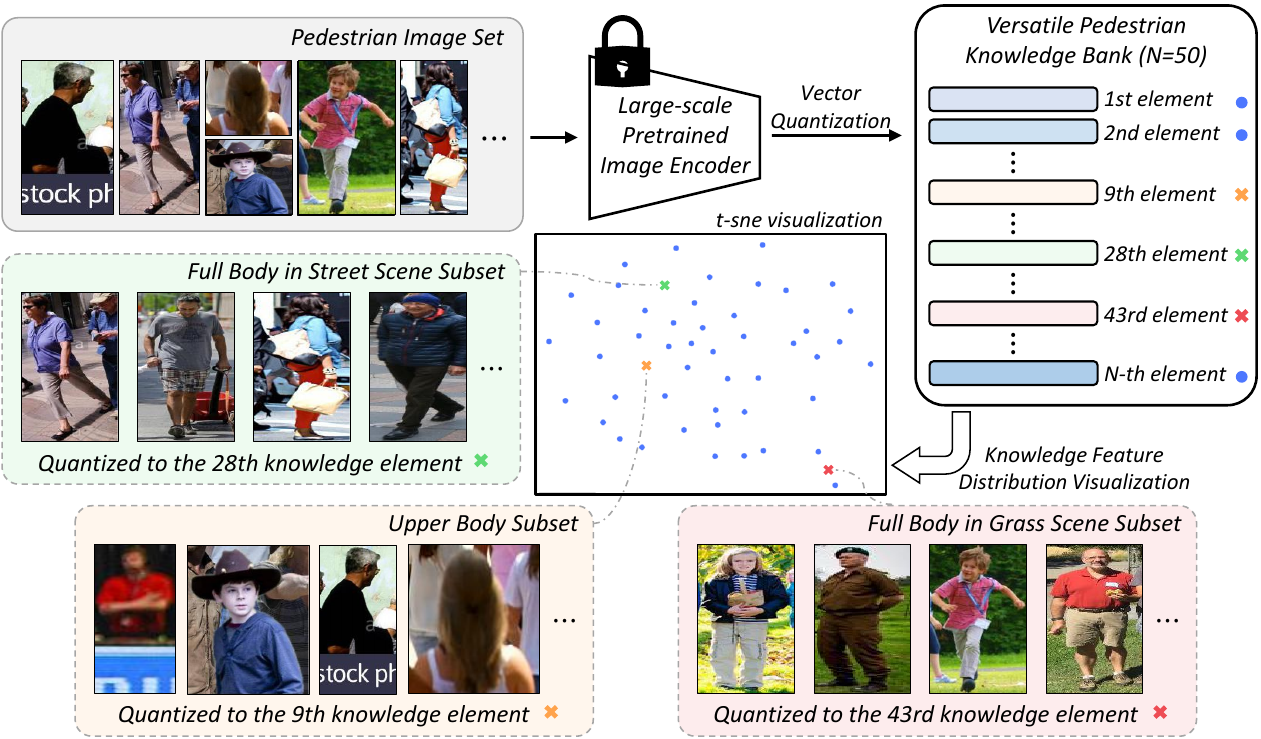}
    \caption{The visualization analysis of semantics in the knowledge bank. We analyze which types of pedestrians are quantized together, and then we visualize the distribution of knowledge features using t-sne. Orange, green, and red $\times$ marks denote the 9th, 28th, and 43rd knowledge elements, respectively, while blue circles are for the others.}
	\label{fig4}
\end{figure*}

In this subsection, we analyze semantics in versatile pedestrian knowledge bank. To this end, we inspect which input pedestrian instances are quantized into which element of the knowledge bank. Figure \ref{fig4} illustrates the feature visualization of versatile pedestrian knowledge bank and analysis of semantics in the bank. To analyze the semantics, we feed pedestrian image sets into CLIP image encoder and conduct vector quantization. Then each pedestrian embedding is mapped into the closest element respectively among $N$ knowledge elements in the bank. So, we can obtain $N$ groups of pedestrians, and we figure out which types of pedestrians mostly appear in each group. We visualize the feature distribution of $N$ knowledge features using t-sne, and we show three examples describing which kinds of pedestrians are quantized into them. Orange, green, and red $\times$ marks denote the 9th, 28th, and 43rd knowledge elements, respectively, while blue circles represent the other elements. As described in the figure, it is observed that the upper body of pedestrians tends to be mapped into the 9th knowledge element $\boldsymbol{f^9_k}$. Furthermore, pedestrians in each of street and grass scenes are quantized into the 28th and 43rd knowledge element, $\boldsymbol{f^{28}_k}$ and $\boldsymbol{f^{43}_k}$, respectively. It means that $\boldsymbol{f^{28}_k}$ and $\boldsymbol{f^{43}_k}$ represent the semantic information of pedestrians in specific scenes. Especially, the street scenes are the primary domain emerging in the driving scene data. Based on the analysis, the versatile pedestrian knowledge bank can contain various semantic information for representing pedestrians and, so that it can help to perform robust pedestrian detection in diverse scene data.

%%%%%%%%%%%%%%%%%%%%%%%%%%%%%%%%%%% Table 3 %%%%%%%%%%%%%%%%%%%%%%%%%%%%%%%%%%%
% \rule{0pt}{9pt}   
\begin{table}[t]
	\caption{The effectiveness of our method on surveillance and driving scenes (WiderPedestrian \cite{wider}). `\# Queries' means the number of queries, and we adopt average precision (AP) as evaluation metric. The higher AP means the better performance.}
	\centering
	\begin{center}
		\renewcommand{\tabcolsep}{4.0mm}
		\resizebox{0.99\linewidth}{!}
		{\begin{tabular}{ccc}
			\Xhline{3\arrayrulewidth}
            \rule{0pt}{11pt}
			\bf Method   & \bf \# Queries & \bf AP ($\uparrow$) \\ \hline
            \rule{0pt}{11pt}
            D-DETR (arXiv'20) \cite{deformable} & 500 & 74.4 \\
            Cascade R-CNN (CVPR'18) \cite{cascade} & - & 74.8 \\
            Sparse R-CNN (CVPR'21) \cite{sparse} & 500 & 76.0 \\                
            E2EDET (CVPR'22) \cite{progressive} & 500 & 77.2 \\ \hline
            \rule{0pt}{11pt}
            \bf Ours & \bf 500 & \bf 77.8 \\ \Xhline{3\arrayrulewidth}
		\end{tabular}
		}
	\end{center}
	\label{tab3} 
\end{table}
%%%%%%%%%%%%%%%%%%%%%%%%%%%%%%%%%%% Table 3 %%%%%%%%%%%%%%%%%%%%%%%%%%%%%%%%%%%

%%%%%%%%%%%%%%%%%%%%%%%%%%%%%%%%% figure 4 %%%%%%%%%%%%%%%%%%%%%%%%%%%%%%%%%

%%%%%%%%%%%%%%%%%%%%%%%%%%%%%%%%% figure 4 %%%%%%%%%%%%%%%%%%%%%%%%%%%%%%%%%

\subsection{Ablation Studies}
We conduct two ablation studies with respect to 1) varying number of features in versatile pedestrian knowledge bank ($N$), 2) the effects of learnable hint $\boldsymbol{f_h}$, and 3) the data variation used for constructing the knowledge bank. The first and second ablation studies are performed on CrowdHuman \cite{crowdhuman} using Sparse R-CNN baseline \cite{sparse}.

\subsubsection{Number of Knowledge Features}
Table \ref{tab4} shows the experimental results with varying number of knowledge features in the bank. We change the number of knowledge features in the bank $N$ with 10, 20, 50 (\textit{default}), 100, and 200. As shown in the table, compared to the baseline, the knowledge bank with only 10 elements can boost the detection performance significantly. And also, when it increases to 200, the detection performances are saturated showing consistent improvements. From the experiments, we could conjecture that the versatile pedestrian knowledge bank is insensitive with the number of elements to obtain robust pedestrian detection performances.

%%%%%%%%%%%%%%%%%%%%%%%%%%%%%%%%%%% Table 4 %%%%%%%%%%%%%%%%%%%%%%%%%%%%%%%%%%%
\begin{table}[t]
	\caption{The ablation study with varying number of knowledge features in the bank.}
	\centering
	\begin{center}
		\renewcommand{\tabcolsep}{6.0mm}
		\resizebox{0.99\linewidth}{!}
		{\begin{tabular}{ccc}
			\Xhline{3\arrayrulewidth}
            \rule{0pt}{11pt}
			\bf Number of Features ($N$) & \bf AP ($\uparrow$) & \bf MR$^{-2}$ ($\downarrow$) \\ \hline
            \rule{0pt}{11pt}
            Sparse R-CNN \cite{sparse} & 90.7 & 44.7 \\ \hdashline
            10 & 92.2 & 42.2 \\
            20 & 92.4 & 42.1 \\
            \bf 50 & \bf 92.4 & \bf 42.1 \\
            100 & 92.3 & 42.5 \\
            200 & 92.3 & 42.5 \\ \Xhline{3\arrayrulewidth}
		\end{tabular}
		}
	\end{center}
	\label{tab4}
\end{table}
%%%%%%%%%%%%%%%%%%%%%%%%%%%%%%%%%%% Table 4 %%%%%%%%%%%%%%%%%%%%%%%%%%%%%%%%%%%

%%%%%%%%%%%%%%%%%%%%%%%%%%%%%%%%%%% Table 5 %%%%%%%%%%%%%%%%%%%%%%%%%%%%%%%%%%%
\begin{table}[t]
	\caption{The ablation study to show the effectiveness of $\boldsymbol{f_h}$.}
	\centering
	\begin{center}
		\renewcommand{\tabcolsep}{4.0mm}
		\resizebox{0.99\linewidth}{!}
		{\begin{tabular}{ccc}
			\Xhline{3\arrayrulewidth}
                \rule{0pt}{11pt}
			\bf Learnable Representation Hint $\boldsymbol{f_h}$ & \bf AP ($\uparrow$) & \bf MR$^{-2}$ ($\downarrow$) \\ \hline
                \rule{0pt}{11pt}
                Sparse R-CNN \cite{sparse} & 90.7 & 44.7 \\ \hdashline
                \xmark & 92.1 & 42.2 \\
                \cmark & \bf 92.4 & \bf 42.1 \\ \Xhline{3\arrayrulewidth}
		\end{tabular}
		}
	\end{center}
	\label{tab5}
\end{table}
%%%%%%%%%%%%%%%%%%%%%%%%%%%%%%%%%%% Table 5 %%%%%%%%%%%%%%%%%%%%%%%%%%%%%%%%%%%

\subsubsection{Effects of Learnable Representation Hint}
Since the representations from a large-scale pretrained model could be suboptimal, we further guide them to be more task-relevant. Therefore, as described in Section. \ref{sect3a}, we place the learnable representation hint $\boldsymbol{f_h}$ to augment and guide the quantized pedestrian representation $\boldsymbol{f_q}$ to be more distinguishable from various backgrounds. So, instead of using $\boldsymbol{f_k} = \boldsymbol{f_q} \oplus \boldsymbol{f_h}$, we try to leverage $\boldsymbol{f_q}$ directly. Table \ref{tab5} shows the performance changes depending on the learnable representation hint $\boldsymbol{f_h}$. The \xmark \ mark denotes that a pedestrian detection framework utilizes $\boldsymbol{f_q}$ directly, and \cmark \ mark means that the framework exploits $\boldsymbol{f_k} = \boldsymbol{f_q} \oplus \boldsymbol{f_h}$. As described in the table, although $\boldsymbol{f_q}$ obtains a performance gain, $\boldsymbol{f_h}$ can bring another performance improvement. The experimental could show that it would be helpful to guide the features to be more suitable for the downstream pedestrian task, rather than directly using the representations directly.

%%%%%%%%%%%%%%%%%%%%%%%%%%%%%%%%% figure 5 %%%%%%%%%%%%%%%%%%%%%%%%%%%%%%%%%
\begin{figure*}[t]
\centering
    \includegraphics[width=0.999\textwidth]{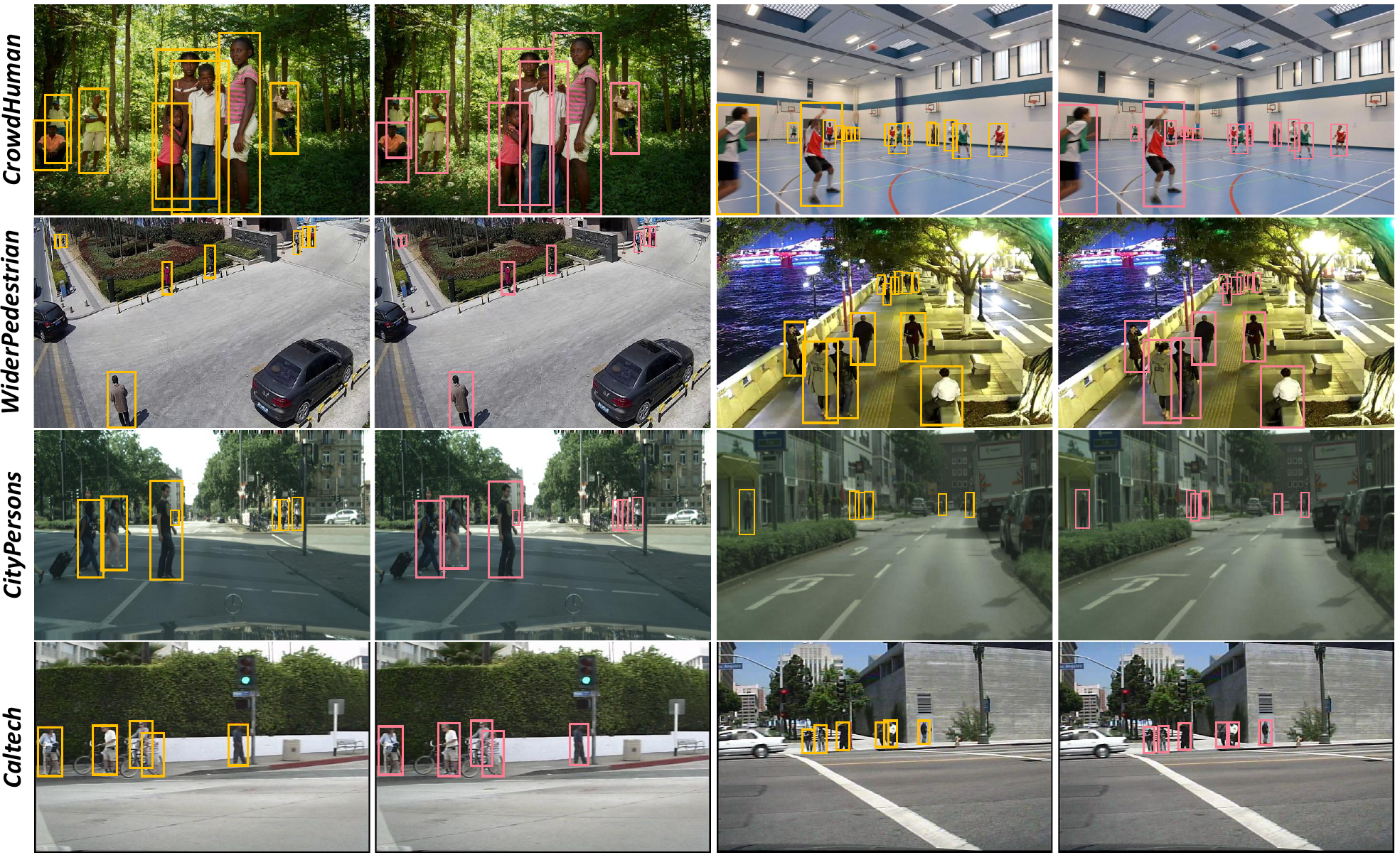}
    \caption{The visualization of detection results on diverse scenes. The \textit{yellow} and \textit{red} boxes mean ground-truth and prediction bounding boxes, respectively. The proposed method performs properly on general indoor/outdoor, surveillance, and driving environments. The images are zoomed in for the better visualization.}
	\label{fig5}
\end{figure*}
%%%%%%%%%%%%%%%%%%%%%%%%%%%%%%%%% figure 5 %%%%%%%%%%%%%%%%%%%%%%%%%%%%%%%%%

\subsubsection{Data Variation for Obtaining Knowledge Features}
When constructing the knowledge bank, we primarily adopt CrowdHuman (CH) for all the experiments. However, to observe the effect of data, we build the knowledge bank with other data: a combination of EuroCityPersons (ECP), CityPersons (CP), and Caltech (C) and a combination of four datasets (CH, ECP, CP, C). Note that we adopt Sparse R-CNN as the baseline detection framework, and we evaluate it on CrowdHuman and CityPersons datasets. Table \ref{tab6} describes the experimental results, and it shows consistent performance improvements when the proposed method is applied. However, as shown in the last row, even though we additionally integrate ECP, CP, and C with CH, the performance gains seem relatively small. We conjecture the reason as follows: Since CH includes numerous pedestrian instances from web-crawled images, it can provide diverse pedestrian knowledge including street scenes. On the other hand, the others could provide relatively less diverse pedestrian knowledge, because these datasets were primarily collected from driving street scenes.

%%%%%%%%%%%%%%%%%%%%%%%%%%%%%%%%%%% Table 6 %%%%%%%%%%%%%%%%%%%%%%%%%%%%%%%%%%%
\begin{table}[t]
	\caption{The ablation study of data variation for constructing the bank (CH: CrowdHuman, ECP: EuroCityPersons, CP: CityPersons, C: Caltech). We evaluate it on CrowdHuman and CityPersons, respectively.}
	\centering
	\begin{center}
		\renewcommand{\tabcolsep}{1.5mm}
		\resizebox{0.99\linewidth}{!}
		{\begin{tabular}{ccc}
			\Xhline{3\arrayrulewidth}
            \rule{0pt}{12pt}
			\bf Dataset & \bf CrowdHuman (AP $\uparrow$)  & \bf CityPersons (MR$^{-2}$ $\downarrow$) \\ \hline
            \rule{0pt}{12pt}
            Sparse R-CNN \cite{sparse} & 90.7 & 10.0 \\ \hdashline
            \bf CH & 92.4 & 7.6 \\
            \bf ECP$+$CP$+$C & 92.4 & \bf 7.5 \\
            \bf CH$+$ECP$+$CP$+$C & \bf 92.5 & \bf 7.5 \\ \Xhline{3\arrayrulewidth}
		\end{tabular}
		}
	\end{center}
	\label{tab6}
\end{table}
%%%%%%%%%%%%%%%%%%%%%%%%%%%%%%%%%%% Table 6 %%%%%%%%%%%%%%%%%%%%%%%%%%%%%%%%%%%

\subsection{Detection Result Visualization}
Figure \ref{fig5} shows the qualitative visualization results of the proposed method on diverse scene data. There are pairs of detection result images and ground-truth images. The red boxes denote the predicted detection boxes, and the yellow boxes mean the ground-truth boxes, respectively. The first row shows the examples from CrowdHuman benchmark \cite{crowdhuman} which consists of web-crawled images including various outdoor and indoor scenes (\eg forest and gym). The second row describes the results on WiderPedestrian dataset \cite{wider} which is mainly collected in surveillance and driving scenes. Lastly, the detection results of CityPersons \cite{city} and Caltech \cite{caltech} are shown in the third and fourth rows. These datasets are composed of driving scene images in urban city. The examples above show that the proposed method can be well adopted in various scene data qualitatively, even though versatile pedestrian knowledge bank is constructed with the training data of CrowdHuman only.

%%%%%%%%%%%%%%%%%%%%%%%%%%%%%%%%% figure 6 %%%%%%%%%%%%%%%%%%%%%%%%%%%%%%%%%
\begin{figure*}[t]
\centering
	\includegraphics[width=0.999\textwidth]{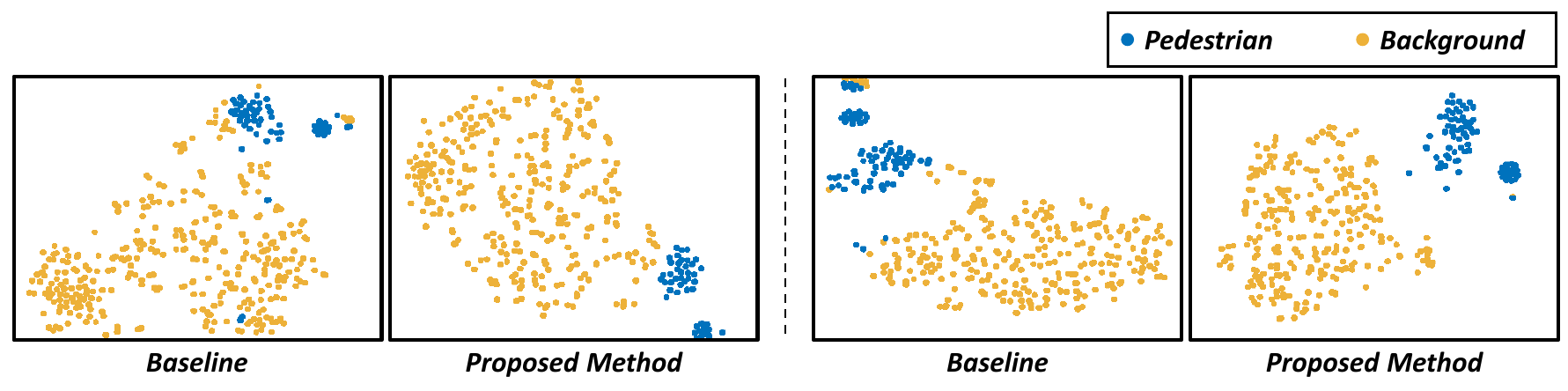}
    \caption{Two examples of t-SNE visualization (\textit{left} and \textit{right}). Compared to the baseline, the proposed method helps the pedestrian features to be more aggregated and separated from the background features easily.}
	\label{fig6}
\end{figure*}
%%%%%%%%%%%%%%%%%%%%%%%%%%%%%%%%% figure 6 %%%%%%%%%%%%%%%%%%%%%%%%%%%%%%%%%

\subsection{Feature Distribution Visualization}
In this subsection, we analyze the changes of feature distribution when applying the proposed method. We adopt Sparse R-CNN \cite{sparse} as baseline, and we use CrowdHuman \cite{crowdhuman}. To analyze the feature distribution, we separate pedestrian and background features depending on intersection-of-union (IoU) threshold. Following \cite{iou2}, we assign the features which have higher IoU than 0.7 with any ground-truth boxes as pedestrian features, and we consider the background features which have lower IoU than 0.3. Figure \ref{fig6} illustrates two examples of t-sne feature distribution visualization. The blue and yellow dots represent the feature distributions of pedestrians and backgrounds, respectively. Please note that, two examples mean the feature distributions obtained from two raw images, and each of them comes from a single raw image. As shown in the distributions obtained from the baseline, pedestrian and background features look difficult to be separated. However, when the proposed method is employed, pedestrian features tend to be more distinctive from background features. In other words, versatile pedestrian knowledge bank can complement the pedestrian features within a pedestrian detection framework and enhance their feature quality to be more distinguishable.

\section{Discussion}
\subsection{Function of $\boldsymbol{f_h}$}
The function of learnable representation hint $\boldsymbol{f_h}$ is to transform the quantized representation $\boldsymbol{f_q}$, obtained from a large-scale pretrained model (\textit{i.e.,} CLIP), to be better aligned with our purpose, pedestrian detection. It is theoretically based on that: large-scale pretrained models usually lack the capability of extracting fine-grained characteristics. So, their representations are suboptimal on downstream tasks and harm transferring ability because of semantic gap between training objectives/distributions \cite{noisy1, vt-clip, domain-aligned}. Therefore, rather than fine-tuning the entire large model and not directly using $\boldsymbol{f_q}$, we place learnable $\boldsymbol{f_h}$ and guide it by pedestrian classification. Then we obtain $\boldsymbol{f_k}$ by adjusting $\boldsymbol{f_q}$ to be compatible with pedestrian detection.

\subsection{Limitation}
The weakness of the proposed method is that it is required to prepare the knowledge bank in advance, which means we could obtain robust detection frameworks not in an end-to-end manner. However, it is also feasible to obtain the representations without data bias for robust detection in an end-to-end manner. Therefore, in future work, it seems interesting to explore the way to obtain and utilize such representations simultaneously.

\section{Conclusion}
In this paper, we presented an innovative method to construct versatile pedestrian knowledge bank. Based on the generalized knowledge from CLIP image encoder, we curate it to obtain the most representative and task-compatible knowledge via vector quantization. We constructed versatile pedestrian knowledge bank by storing task-compatible knowledge. Then we leveraged the knowledge bank into a pedestrian detection framework and complemented pedestrian features within it. Through comprehensive experiments, we validated the effectiveness and versatility. While the proposed method achieved state-of-the-art performances on four public pedestrian detection datasets, we showed that it can be adopted in various detection frameworks, such as two stage based, region proposal based, and query based detectors. As corroborated in the experiments, once the knowledge bank is constructed, it is applicable to a variety of detection frameworks and scene data. Even though our method requires a two-step procedure constructing the knowledge bank and integrating it with a detection framework, it would be an interesting future work to build an end-to-end framework while achieving similar benefits. We believe that our method presents a novel way to obtain versatile representations for a wide range of studies in the multi-domain fields.

\section{Acknowledgement}
We would like to thank \textit{Jongwook Han} for helping us to conduct the experiments. This work was supported by IITP grant funded by the Korea government (MSIT) (No. 2020-0-00004, Development of Previsional Intelligence based on Long-Term Visual Memory Network).

% \clearpage
\bibliographystyle{elsarticle-num} 
\bibliography{elsarticle}

\begin{thebibliography}{10}
\expandafter\ifx\csname url\endcsname\relax
  \def\url#1{\texttt{#1}}\fi
\expandafter\ifx\csname urlprefix\endcsname\relax\def\urlprefix{URL }\fi
\expandafter\ifx\csname href\endcsname\relax
  \def\href#1#2{#2} \def\path#1{#1}\fi

\bibitem{pr_ped1}
X.~Gao, Y.~Xiong, G.~Zhang, H.~Deng, K.~Kou, Exploiting key points supervision and grouped feature fusion for multiview pedestrian detection, Pattern Recognition 131 (2022) 108866.

\bibitem{pr_ped2}
Z.~Guo, W.~Liao, Y.~Xiao, P.~Veelaert, W.~Philips, Weak segmentation supervised deep neural networks for pedestrian detection, Pattern Recognition 119 (2021) 108063.

\bibitem{lowresolution}
Y.-F. Liu, J.-M. Guo, C.-H. Chang, Low resolution pedestrian detection using light robust features and hierarchical system, Pattern Recognition 47~(4) (2014) 1616--1625.

\bibitem{liu2023weakly}
M.~Liu, Y.~Bian, Q.~Liu, X.~Wang, Y.~Wang, Weakly supervised tracklet association learning with video labels for person re-identification, IEEE Transactions on Pattern Analysis and Machine Intelligence (2023).

\bibitem{liu2024two}
M.~Liu, F.~Wang, X.~Wang, Y.~Wang, A.~K. Roy-Chowdhury, A two-stage noise-tolerant paradigm for label corrupted person re-identification, IEEE Transactions on Pattern Analysis and Machine Intelligence (2024).

\bibitem{pd11}
Y.~Tang, M.~Liu, B.~Li, Y.~Wang, W.~Ouyang, Otp-nms: Towards optimal threshold prediction of nms for crowded pedestrian detection, IEEE Transactions on Image Processing (2023).

\bibitem{wang2023prototype}
Q.~Wang, W.~Zhang, W.~Yang, C.~Xu, Z.~Cui, Prototype-guided instance matching for multiple pedestrian tracking, Neurocomputing 538 (2023) 126207.

\bibitem{pr_ped3}
Y.~Jin, Y.~Zhang, Y.~Cen, Y.~Li, V.~Mladenovic, V.~Voronin, Pedestrian detection with super-resolution reconstruction for low-quality image, Pattern Recognition 115 (2021) 107846.

\bibitem{kim2021robust}
J.~U. Kim, S.~Park, Y.~M. Ro, Robust small-scale pedestrian detection with cued recall via memory learning, in: Proceedings of the IEEE/CVF international conference on computer vision, 2021, pp. 3050--3059.

\bibitem{ped4}
S.~Liu, D.~Huang, Y.~Wang, Adaptive nms: Refining pedestrian detection in a crowd, in: Proceedings of the IEEE/CVF Conference on Computer Vision and Pattern Recognition, 2019, pp. 6459--6468.

\bibitem{iterdet}
D.~Rukhovich, K.~Sofiiuk, D.~Galeev, O.~Barinova, A.~Konushin, Iterdet: Iterative scheme for object detection in crowded environments, in: Joint IAPR International Workshops on Statistical Techniques in Pattern Recognition (SPR) and Structural and Syntactic Pattern Recognition (SSPR), 2021, pp. 344--354.

\bibitem{progressive}
A.~Zheng, Y.~Zhang, X.~Zhang, X.~Qi, J.~Sun, Progressive end-to-end object detection in crowded scenes, in: Proceedings of the IEEE/CVF Conference on Computer Vision and Pattern Recognition, 2022, pp. 857--866.

\bibitem{elephant}
I.~Hasan, S.~Liao, J.~Li, S.~U. Akram, L.~Shao, Generalizable pedestrian detection: The elephant in the room, in: Proceedings of the IEEE/CVF Conference on Computer Vision and Pattern Recognition, 2021, pp. 11328--11337.

\bibitem{clip}
A.~Radford, J.~W. Kim, C.~Hallacy, A.~Ramesh, G.~Goh, S.~Agarwal, G.~Sastry, A.~Askell, P.~Mishkin, J.~Clark, et~al., Learning transferable visual models from natural language supervision, in: International Conference on Machine Learning, PMLR, 2021, pp. 8748--8763.

\bibitem{vector1}
K.~Lakhotia, E.~Kharitonov, W.-N. Hsu, Y.~Adi, A.~Polyak, B.~Bolte, T.-A. Nguyen, J.~Copet, A.~Baevski, A.~Mohamed, et~al., On generative spoken language modeling from raw audio, Transactions of the Association for Computational Linguistics 9 (2021) 1336--1354.

\bibitem{noisy1}
S.~Petryk, L.~Dunlap, K.~Nasseri, J.~Gonzalez, T.~Darrell, A.~Rohrbach, On guiding visual attention with language specification, in: Proceedings of the IEEE/CVF Conference on Computer Vision and Pattern Recognition, 2022, pp. 18092--18102.

\bibitem{vt-clip}
L.~Qiu, R.~Zhang, Z.~Guo, Z.~Zeng, Y.~Li, G.~Zhang, Vt-clip: Enhancing vision-language models with visual-guided texts, arXiv preprint arXiv:2112.02399 (2021).

\bibitem{cascade}
Z.~Cai, N.~Vasconcelos, Cascade r-cnn: Delving into high quality object detection, in: Proceedings of the IEEE/CVF Conference on Computer Vision and Pattern Recognition, 2018, pp. 6154--6162.

\bibitem{sparse}
P.~Sun, R.~Zhang, Y.~Jiang, T.~Kong, C.~Xu, W.~Zhan, M.~Tomizuka, L.~Li, Z.~Yuan, C.~Wang, et~al., Sparse r-cnn: End-to-end object detection with learnable proposals, in: Proceedings of the IEEE/CVF Conference on Computer Vision and Pattern Recognition, 2021, pp. 14454--14463.

\bibitem{deformable}
X.~Zhu, W.~Su, L.~Lu, B.~Li, X.~Wang, J.~Dai, Deformable detr: Deformable transformers for end-to-end object detection, arXiv preprint arXiv:2010.04159 (2020).

\bibitem{bodla}
N.~Bodla, B.~Singh, R.~Chellappa, L.~S. Davis, Soft-nms--improving object detection with one line of code, in: Proceedings of the IEEE International Conference on Computer Vision, 2017, pp. 5561--5569.

\bibitem{ped3}
X.~Huang, Z.~Ge, Z.~Jie, O.~Yoshie, Nms by representative region: Towards crowded pedestrian detection by proposal pairing, in: Proceedings of the IEEE/CVF Conference on Computer Vision and Pattern Recognition, 2020, pp. 10750--10759.

\bibitem{ped2}
Y.~Zhang, H.~He, J.~Li, Y.~Li, J.~See, W.~Lin, Variational pedestrian detection, in: Proceedings of the IEEE/CVF Conference on Computer Vision and Pattern Recognition, 2021, pp. 11622--11631.

\bibitem{attribute}
J.~Zhang, L.~Lin, J.~Zhu, Y.~Li, Y.-c. Chen, Y.~Hu, S.~C. Hoi, Attribute-aware pedestrian detection in a crowd, IEEE Transactions on Multimedia 23 (2020) 3085--3097.

\bibitem{zero6}
K.~Kobs, M.~Steininger, A.~Hotho, Indirect: Language-guided zero-shot deep metric learning for images, in: Proceedings of the IEEE/CVF Winter Conference on Applications of Computer Vision, 2023, pp. 1063--1072.

\bibitem{zero7}
A.~Sanghi, H.~Chu, J.~G. Lambourne, Y.~Wang, C.-Y. Cheng, M.~Fumero, K.~R. Malekshan, Clip-forge: Towards zero-shot text-to-shape generation, in: Proceedings of the IEEE/CVF Conference on Computer Vision and Pattern Recognition, 2022, pp. 18603--18613.

\bibitem{zero9}
X.~Gu, T.-Y. Lin, W.~Kuo, Y.~Cui, Open-vocabulary object detection via vision and language knowledge distillation, arXiv preprint arXiv:2104.13921 (2021).

\bibitem{unihcp}
Y.~Ci, Y.~Wang, M.~Chen, S.~Tang, L.~Bai, F.~Zhu, R.~Zhao, F.~Yu, D.~Qi, W.~Ouyang, Unihcp: A unified model for human-centric perceptions, in: CVPR, 2023, pp. 17840--17852.

\bibitem{crowdhuman}
S.~Shao, Z.~Zhao, B.~Li, T.~Xiao, G.~Yu, X.~Zhang, J.~Sun, Crowdhuman: A benchmark for detecting human in a crowd, arXiv preprint arXiv:1805.00123 (2018).

\bibitem{wider}
C.~C. Loy, D.~Lin, W.~Ouyang, Y.~Xiong, S.~Yang, Q.~Huang, D.~Zhou, W.~Xia, Q.~Li, P.~Luo, et~al., Wider face and pedestrian challenge 2018: Methods and results, arXiv preprint arXiv:1902.06854 (2019).

\bibitem{city}
S.~Zhang, R.~Benenson, B.~Schiele, Citypersons: A diverse dataset for pedestrian detection, in: Proceedings of the IEEE/CVF Conference on Computer Vision and Pattern Recognition, 2017, pp. 3213--3221.

\bibitem{caltech}
P.~Dollar, C.~Wojek, B.~Schiele, P.~Perona, Pedestrian detection: An evaluation of the state of the art, IEEE Transactions on Pattern Analysis and Machine Intelligence 34~(4) (2011) 743--761.

\bibitem{fpn}
T.-Y. Lin, P.~Doll{\'a}r, R.~Girshick, K.~He, B.~Hariharan, S.~Belongie, Feature pyramid networks for object detection, in: Proceedings of the IEEE/CVF Conference on Computer Vision and Pattern Recognition, 2017, pp. 2117--2125.

\bibitem{cvpr23_1}
X.~Song, B.~Chen, P.~Li, J.-Y. He, B.~Wang, Y.~Geng, X.~Xie, H.~Zhang, Optimal proposal learning for deployable end-to-end pedestrian detection, in: Proceedings of the IEEE/CVF Conference on Computer Vision and Pattern Recognition, 2023, pp. 3250--3260.

\bibitem{ped5}
J.~Wang, C.~Zhao, Z.~Huo, Y.~Qiao, H.~Sima, High quality proposal feature generation for crowded pedestrian detection, Pattern Recognition 128 (2022) 108605.

\bibitem{alfnet}
W.~Liu, S.~Liao, W.~Hu, X.~Liang, X.~Chen, Learning efficient single-stage pedestrian detectors by asymptotic localization fitting, in: Proceedings of the European Conference on Computer Vision, 2018, pp. 618--634.

\bibitem{csp}
W.~Liu, I.~Hasan, S.~Liao, Center and scale prediction: Anchor-free approach for pedestrian and face detection, Pattern Recognition 135 (2023) 109071.

\bibitem{bgc}
J.~Li, S.~Liao, H.~Jiang, L.~Shao, Box guided convolution for pedestrian detection, in: Proceedings of the 28th ACM International Conference on Multimedia, 2020, pp. 1615--1624.

\bibitem{iou2}
Y.~Yang, F.~Wei, M.~Shi, G.~Li, Restoring negative information in few-shot object detection, Advances in Neural Information Processing Systems 33 (2020) 3521--3532.

\bibitem{domain-aligned}
M.~W. Gondal, J.~Gast, I.~A. Ruiz, R.~Droste, T.~Macri, S.~Kumar, L.~Staudigl, Domain aligned clip for few-shot classification, in: Proceedings of the IEEE/CVF Winter Conference on Applications of Computer Vision, 2024, pp. 5721--5730.

\end{thebibliography}

\end{document}